\documentclass[3p,times,procedia]{elsarticle}
\usepackage{ecrc}
\usepackage[bookmarks=false]{hyperref}
\usepackage{booktabs}
    \hypersetup{colorlinks,
      linkcolor=blue,
      citecolor=blue,
      urlcolor=blue}

\usepackage{amsmath,amssymb}
\usepackage{graphicx}
\usepackage{booktabs}
\usepackage{array}
\usepackage{url}
\usepackage{xcolor}
\usepackage{multirow}

\biboptions{sort&compress}

\newcommand{\Lcal}{\mathcal{L}}

\newcommand{\Lfoc}{\Lcal_{\text{Focal}}}
\newcommand{\Lali}{\Lcal_{\text{Align}}}
\newcommand{\Latt}{\Lcal_{\text{Atten}}}

\volume{00}

\firstpage{1}

\journalname{Procedia Computer Science}

\runauth{Paper-ID 1807}

\jid{procs}

\usepackage{amssymb}
\usepackage{amsthm}

\usepackage[figuresright]{rotating}

\begin{document}
\begin{frontmatter}



\dochead{5th International Conference on Machine Learning and Data Engineering}

\title{\textsc{PRCA-Net}: A Physics-Restored Cross-Attention Network for Breast Ultrasound Lesion Classification}

\author[a]{Nikunja Bihari Kar}
\author[b]{Ishita Jena}
\author[b]{Arpita Mishra}
\author[a]{Ankit Jha\corref{cor1}}

\address[a]{The LNM Institute of Information Technology, Jaipur, India}
\address[b]{Independent Researcher}


\begin{abstract}
Breast ultrasound (BUS) is widely used for lesion assessment, but speckle,
depth-dependent attenuation, and acoustic shadowing can hinder automated
classification. We propose \textsc{PRCA-Net}: Physics-Restored Cross-Attention
Network, a dual-stream framework that learns from physics-based augmented and
restored views of a B-mode image. The restored view is generated through adaptive
despeckling and attenuation compensation. Independent EfficientNet-B0 encoders
extract features from both views, which are integrated using bidirectional
cross-attention. To preserve cross-stream correspondence without forcing the
representations to become redundant, we introduce a banded cosine alignment loss
on corresponding spatial tokens. We further regularize the fused representation
with a depth-wise attenuation-consistency loss. Experiments on BUSI, BUS-BRA, and
UDIAT demonstrates competitive performance, and ablation studies validate the
contributions of the proposed alignment and attenuation-consistency objectives.
\end{abstract}

\begin{keyword}
Breast ultrasound; Lesion classification; Physics-guided deep learning; Cross-attention; Dual-stream framework;




\end{keyword}
\cortext[cor1]{Corresponding author.}
\end{frontmatter}

\email{ankitjha16@gmail.com}

\vspace*{-6pt}

\section{Introduction}

Breast cancer is a major global health burden, with approximately 2.3
million new diagnoses each year~\cite{bray2024global}. Breast ultrasound (BUS) is
widely used for lesion assessment due to its non-ionizing nature, relatively low cost, portability and suitability for real time imaging. It is particularly
useful for dense breasts, where mammography can be less sensitive~\cite{berg2008combined}.
Because of these benefits, BUS is a crucial computer-aided diagnostic modality, particularly in situations where professional interpretation is not always accessible.

Automated BUS classification remains challenging because a B-mode image is shaped
by the ultrasound acquisition process as well as by tissue morphology. Coherent
interference among sub-resolution scatterers produces speckle, whose amplitude is
Rayleigh distributed in the fully developed case~\cite{burckhardt1978speckle}.
Speckle can obscure lesion texture, but it may also retain information related to
tissue microstructure. Ultrasound energy further decreases with depth,
approximately following $I(z)=I_0e^{-\alpha z}$~\cite{szabo2013diagnostic},
which leads to depth-dependent intensity variation. Attenuating lesions may also
produce posterior acoustic shadowing, a sonographic finding associated with
malignancy~\cite{stavros1995solid}. Consequently, acquisition effects can act as
both nuisance variation and potentially useful diagnostic evidence.

Most BUS classifiers adapt image-recognition backbones and optimize them with
classification supervision. Such models are not explicitly encouraged to separate
tissue-related appearance from acquisition-dependent effects. Clinical-prior
methods address this limitation from a different direction, for example through
BI-RADS concept bottlenecks~\cite{bunnell2024concept} or guideline-constrained
prototypes~\cite{yan2024domain}. These methods do not explicitly replicate the physical effects that shape the input image, but they do enhance the clinical structure of the decision-making process. Other techniques use wavelet-based processing or adaptive filtering to suppress speckle prior to classification~\cite{lee1980digital,onjun2026wavelet}.
Although this kind of preprocessing can enhance picture quality, it restricts the classifier to a single representation and may suppress both undesirable artifacts and valuable fine-scale data. Recent dual-branch approaches show that complementary views can improve BUS
classification. DBL-Net integrates spatial and frequency-domain features~\cite{zhu2024dblnet}, DSDNet integrates a segmentation-derived prior~\cite{jiang2026dsdnet}, and HADS-Net combines texture features with Sobel edge maps and physics-informed augmentation~\cite{mbonu2026hadsnet}. These techniques make use of many complimentary information types, such as task-derived priors, frequency content, and boundaries. They control the connection between their learnt representations, but they do not specifically integrate a second physics-guided perspective with a restoration-derived view.

We propose a dual-stream architecture called \textsc{PRCA-Net} (Physics-Restored Cross-Attention Network), which learns from both physics-augmented and physics-restored views of the same B-mode image. The restored view is generated
through adaptive despeckling and depth-dependent attenuation compensation. The model shares complementary information across relevant spatial regions since the two views are encoded individually and combined via bidirectional cross-attention. This design raises an additional issue that the two
representations should remain sufficiently related for cross-attention, but they
should not become identical. In our context, unrestricted agreement can lower the value of the second stream. Standard representation-alignment aims encourage agreement between paired views~\cite{zbontar2021barlow,chen2021exploring}. In order to preserve the similarity between matching tokens inside a specified interval, we use a banded cosine alignment loss. This formulation is related
to the observation that useful views share partial, rather than identical,
information~\cite{tian2020makes}. Furthermore, we apply a depth-wise attenuation-consistency requirement to the fused feature representation. Rather than prescribing a fixed attenuation rate,
the constraint encourages a smooth, approximately log-linear feature-energy
profile along the depth direction. The final model combines class-weighted focal
loss~\cite{lin2017focal}, banded alignment, and attenuation consistency. We
evaluate \textsc{PRCA-Net} on BUSI, BUS-BRA, and UDIAT using a unified
class-stratified 70/15/15 train/validation/test protocol with five random seeds
and validation-selected decision thresholds. We summarize our contributions as follows:
\begin{itemize}
    \item We propose \textsc{PRCA-Net}, a dual-stream framework for breast
    ultrasound classification that jointly learns from physics-augmented and
    physics-restored views of a B-mode image.

    \item We introduce bidirectional cross-attention with banded cosine
    alignment, which promotes correspondence between the two streams while
    preserving view-specific information.

    \item We introduce a depth-wise attenuation-consistency constraint on the
    fused feature representation, incorporating ultrasound-inspired physical
    structure directly into feature learning.

    \item Experimental demonstration of \textsc{PRCA-Net} on BUSI, BUS-BRA, and UDIAT, and provide
    ablation studies of the proposed alignment and attenuation-consistency
    objectives.
\end{itemize}

\section{Related Work}
\label{sec:related}

\paragraph{Transfer learning and clinical priors}
Most breast ultrasound classification methods build on CNNs pretrained on natural
images. ResNet~\cite{he2016deep}, DenseNet~\cite{huang2017densely}, and
EfficientNet~\cite{tan2019efficientnet} are frequently used as backbones, often
with attention modules such as CBAM~\cite{woo2018cbam} to emphasize informative
image regions. Vision Transformers have also been explored for image recognition
and medical imaging applications~\cite{dosovitskiy2021image}. However, BUS
datasets are usually much smaller than the datasets used to pretrain these models,
and ultrasound images vary considerably across scanners, operators, and acquisition
settings. Because of this, a model can rely on texture or intensity patterns without being aware of whether they originate from the imaging process or depict lesion tissue. Another group of methods introduces clinical knowledge into the classifier.
Diagnostic characteristics that radiologists can examine and, if necessary, correct are represented by BI-RADS concept bottlenecks~\cite{bunnell2024concept}. Similarly,
guideline-based prototype methods use clinically meaningful rules to structure the
decision process~\cite{yan2024domain}. These approaches make the prediction more
interpretable and incorporate information that is useful in clinical practice.
However, they mainly focus on diagnostic concepts and decision rules. They do not
explicitly address the ultrasound effects that influence the image before feature
extraction, such as speckle, depth-dependent attenuation, and acoustic shadowing.
Our work complements these directions by introducing ultrasound-specific physical
information at the representation-learning stage.

\paragraph{Multi-branch architectures and speckle handling}
Speckle reduction has long been studied in ultrasound image analysis. Classical
methods such as the Lee filter~\cite{lee1980digital} reduce speckle by using local
image statistics, while more recent methods can suppress high-frequency
components inside the network. For example, wavelet pooling reduces the influence
of high-frequency content during down-sampling~\cite{onjun2026wavelet}. Although
such processing can make an image easier to analyze, it can also remove subtle
texture, weak boundaries, or posterior acoustic patterns that may be useful for
distinguishing benign and malignant lesions. Using only a filtered image therefore
forces the classifier to work with one interpretation of the scan. Recent methods instead use multiple branches to preserve complementary
information. HADS-Net~\cite{mbonu2026hadsnet} employs augmentation influenced by physics and integrates texture features with Sobel edge maps. While DSDNet~\cite{jiang2026dsdnet} utilizes a frozen segmentation before guiding classification, DBL-Net~\cite{zhu2024dblnet} uses spatial and frequency-domain branches.  These approaches show
that additional views can improve BUS classification. Their auxiliary branches, however, concentrate on edges, frequency data, or structure connected to segmentation.
The second branch of our technique uses adaptive despeckling and attenuation correction to produce a physics-restored version of the identical B-mode image.  We keep
both the original and restored views because neither is always preferable. While the restored image might highlight tissue patterns in areas impacted by speckle or attenuation, the raw image can maintain minute features that restoration can decrease. The bidirectional cross-attention allows the proposed model to decide how to use these two views at different spatial locations.

\paragraph{Representation alignment}
A lot of self-supervised learning (SSL) methods learn knowledge from two similar views of an image.
Barlow Twins~\cite{zbontar2021barlow} reduces redundancy across feature dimensions and promotes paired views to generate comparable representations. Additionally, Siamese approaches promote agreement between representations derived from enhanced versions of the same sample~\cite{chen2021exploring}.These techniques are effective when, notwithstanding augmentation, the two perspectives should convey the same semantic meaning. Although they are connected, the two perspectives employed in our approach are not the same. Their traits should be similar as they belong to the same anatomy. They correspond to
the same anatomy, so their features should be comparable. However, because the restoration procedure modifies the appearance of speckles and attenuation, it is anticipated that the restored image would be different from the raw image. Cross-attention might not be able to create meaningful correlation if the two feature streams diverge too much. The second stream adds little new information if they start to resemble each other too much. This concept aligns with the finding that helpful perspectives share some information, but not all of it~\cite{tian2020makes}. In order to maintain the similarity between related tokens within a predetermined range, we employ a banded cosine alignment loss. The lower bound encourages the two streams to remain related,
and the upper bound preserves the complementary information that motivates the
two-stream design. In this way, ultrasound physics influences not only the input
image but also how the two feature representations are related during training.

\section{Methodology}
\label{sec:method}

\subsection{Preliminaries}
\label{sec:overview}

Given a preprocessed B-mode breast ultrasound image
$x_r\in\mathbb{R}^{3\times H\times W}$, \textsc{PRCA-Net} generates two
physics-guided views, as illustrated in Figure~\ref{fig:arch}. The first view,
$x_{pa}$, is obtained through physics-based augmentation and exposes the model to
plausible variations caused by ultrasound acquisition. The second view, $x_{pr}$,
is obtained through physics-based restoration, which reduces speckle variation and
compensates for depth-dependent attenuation. 
The two views are processed by independent encoders. Their representations are
projected to a common embedding space, regularized by a banded alignment loss, and
integrated through bidirectional cross-attention. The attended features are then
concatenated and fused by a convolutional head. During training, the classification
loss, cross-stream alignment loss, and attenuation-consistency loss jointly
optimize the network.

\subsection{Physics-based view generation}
\label{sec:views}

\noindent\textbf{Physics-based augmentation.}
The upper branch in Figure~\ref{fig:arch} applies a physics-based augmentation operator to the input image $x_{pa}=\mathcal{A}_{\mathrm{phys}}(x_r)$, where $\mathcal{A}_{\mathrm{phys}}(\cdot)$ denote the augmentation operator. 
The operator simulates ultrasound-specific appearance variation while retaining
the lesion content and image-level class label. In particular, it may model
changes associated with speckle intensity, depth-dependent gain, and acoustic
shadowing. The augmented view encourages the encoder to learn features that remain
useful under plausible acquisition changes. The exact augmentation parameters are
applied only during training and are described in
Section~\ref{sec:protocol}.

\noindent\textbf{Physics-based restoration.}
The lower branch constructs a restored view $x_{pr} = \mathcal{R}_{phy}(x_{pa})$from the same input image, where $\mathcal{R}_{\mathrm{phys}}(\cdot)$ denote restoration operator. We first
apply adaptive despeckling and then compensate for depth-dependent attenuation.
Both operations preserve image coordinates, so the restored image remains aligned
with the input image.

Let $I(p)$ denote the intensity at pixel $p$. We compute its local mean
$\bar I(p)$ and local variance $\sigma_I^2(p)$ in a $w\times w$ neighborhood,
where $w=5$. A Lee-inspired adaptive filtering rule~\cite{lee1980digital} is used, which produces the despeckled image \vspace{-0.7cm}
\begin{equation}
\hat I(p)
=
\bar I(p)
+
W(p)\bigl[I(p)-\bar I(p)\bigr],
\qquad
W(p)
=
\frac{\sigma_I^2(p)}
{\sigma_I^2(p)+\bar{\sigma}^2},
\label{eq:lee}
\end{equation}
where $\bar{\sigma}^2$ is the mean local variance over the image. Where local variation is large, the filter preserves more of the original intensity and performs stronger smoothing in locally homogenous areas. We then apply a limited gain in the direction of depth. Let $z\in[0,1]$ represent the normalized depth and $u$ represent the horizontal coordinate: \vspace{-0.5cm}
\begin{equation}
I_c(u,z)
=
\min\bigl(e^{\alpha z},G_{\max}\bigr)\hat I(u,z),
\vspace{-0.5cm}
\label{eq:attencomp}
\end{equation}
Although the gain rises with depth, it is limited by $G_{\max}$ to prevent overamplification in low-signal areas. The restored view is defined as $x_{pr}=I_c$ in the generated image. 
The same normalization procedure is applied to $x_{pa}$ and $x_{pr}$ before they
are passed to their respective encoders.

\subsection{Our proposed \textsc{PRCA-Net}}
\label{sec:fusion}

\begin{figure}[!tbh]
\centering
\includegraphics[width=\linewidth]{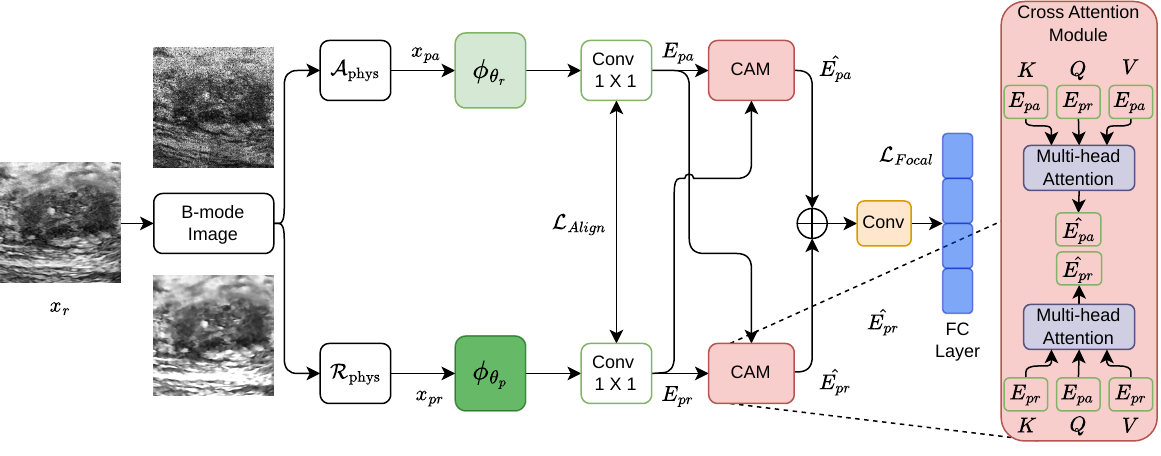}
\caption{The architectural overview of \textsc{PRCA-Net}. A preprocessed B-mode image produces a
raw view and a physics-restored view. Independent encoders extract spatial tokens
from both views. Bidirectional cross-attention exchanges information between the
two streams, while the banded alignment loss $\Lali$ regularizes their
corresponding token representations. A convolutional fusion head preserves spatial
information before global pooling and final classification. }
\label{fig:arch}
\end{figure}

The two views are processed by encoders with identical EfficientNet-B0
architectures but independent learnable parameters. As shown in Figure~\ref{fig:arch}, the
augmentation stream uses $\phi_{\theta_r}$ and the restoration stream uses
$\phi_{\theta_p}$. To make the notation consistent with the view labels, we
denote their projected token representations by $E_{pa}$ and $E_{pr}$: \vspace{-0.5cm}
\begin{equation}
E_{pa}
=
\mathrm{flat}\bigl(
W_{pa}*\phi_{\theta_r}(x_{pa})
\bigr)
\in\mathbb{R}^{N\times d},
\qquad
E_{pr}
=
\mathrm{flat}\bigl(
W_{pr}*\phi_{\theta_p}(x_{pr})
\bigr)
\in\mathbb{R}^{N\times d}, \vspace{-0.5cm}
\label{eq:tokens}
\end{equation}
where $W_{pa}$ and $W_{pr}$ are $1\times1$ convolutional projections,
$d=320$ is the embedding dimension, and $N=hw$ is the number of spatial tokens.
For a $224\times224$ input, EfficientNet-B0 produces a $7\times7$ feature map,
giving $N=49$ tokens. We use bidirectional multi-head cross-attention with $H=8$ heads to exchange
information between the two representations~\cite{vaswani2017attention}. In the
upper direction, $E_{pa}$ provides keys and values, while $E_{pr}$ provides
queries. In the lower direction, $E_{pr}$ provides keys and values, while
$E_{pa}$ provides queries: \vspace{-0.5cm}
\begin{equation}
\hat E_{pa}
=
\mathrm{LN}\bigl(
E_{pa}
+
\mathrm{MHA}(E_{pr},E_{pa},E_{pa})
\bigr),
\qquad
\hat E_{pr}
=
\mathrm{LN}\bigl(
E_{pr}
+
\mathrm{MHA}(E_{pa},E_{pr},E_{pr})
\bigr) \vspace{-0.5cm}
\label{eq:crossattn}
\end{equation}
Here, $\mathrm{MHA}(Q,K,V)$ denotes multi-head attention and
$\mathrm{LN}(\cdot)$ denotes layer normalization. The residual connections retain
view-specific information while allowing each stream to retrieve complementary
evidence from the other view.

The attended token sequences are reshaped into spatial maps
$\hat M_{pa},\hat M_{pr}\in\mathbb{R}^{d\times h\times w}$ and concatenated along
the channel dimension. The convolutional fusion head produces \vspace{-0.5cm}
\begin{equation}
F_f
=
\sigma\left(
\mathrm{BN}\left(
W_2*
\sigma\left(
\mathrm{BN}\left(
W_1*[\hat M_{pa};\hat M_{pr}]
\right)
\right)
\right)
\right),
\label{eq:convfusion} \vspace{-0.5cm}
\end{equation}
where $W_1$ is a $3\times3$ convolution mapping $2d$ channels to $d$ channels,
$W_2$ is a $1\times1$ convolution, $\mathrm{BN}$ denotes batch normalization,
and $\sigma$ is the ReLU activation. The fused feature map $F_f$ is globally
average pooled and passed through dropout and a fully
connected layer to obtain the classification logits.

\subsection{Banded representation alignment}
\label{sec:align}

The two views share the same lesion-level semantics but differ in appearance due
to the augmentation and restoration operations. Their representations should
therefore remain sufficiently related for cross-attention while retaining
view-specific information. We impose this constraint before cross-attention on
the projected tokens $E_{pa}$ and $E_{pr}$, as shown by
$\mathcal{L}_{\mathrm{Align}}$ in Figure~\ref{fig:arch}. For each spatial token $i$, we compute cosine similarity, which is defined as $c_i$ \vspace{-0.7cm}
\begin{equation}
c_i
=
\cos\bigl(E_{pa}^{(i)},E_{pr}^{(i)}\bigr)
\label{eq:cosine}
\end{equation}
We then constrain the similarity to the interval
$[\tau_{\mathrm{lo}},\tau_{\mathrm{hi}}]$ using a banded hinge loss:\vspace{-0.5cm}
\begin{equation}
\Lali
=
\frac{1}{N}\sum_{i=1}^{N}
\left[
\max\bigl(0,\tau_{\mathrm{lo}}-c_i\bigr)
+
\max\bigl(0,c_i-\tau_{\mathrm{hi}}\bigr)
\right],
\vspace{-0.5cm}
\label{eq:align}
\end{equation}
The lower-bound term discourages weak correspondence, whereas the upper-bound
term discourages the two streams from producing redundant features. No penalty is
applied when the token similarity falls within the target interval. For analysis,
we also report the average token similarity $\bar c =\frac{1}{N}\sum_{i=1}^{N}c_i.$

\subsection{Training objective}
\label{sec:loss}

\noindent\textbf{Class-weighted focal loss.}
Since malignant lesions are less frequent than benign lesions, we use
class-weighted focal loss~\cite{lin2017focal} to account the class imbalancing:\vspace{-0.5cm}
\begin{equation}
\vspace{-0.5cm}
\Lfoc
=
-\alpha_y(1-p_y)^\gamma\log p_y,
\label{eq:focal}
\end{equation}
where $p_y$ is the predicted probability of the ground-truth class $y$ and
$\alpha_y$ is the corresponding inverse-frequency class weight. The weights are
computed from the training partition and normalized to have unit mean.

\noindent\textbf{Attenuation-consistency loss.}
The attenuation-consistency loss is applied to the fused representation $F_f$,
rather than to either input view. For each depth row $z$, we compute the average
feature energy in Eq. \ref{eq:featureenergy}.\vspace{-0.5cm}
\begin{equation}
e(z)
=
\mathbb{E}_{c,u}
\left[
F_f(c,u,z)^2
\right], \vspace{-0.5cm}
\label{eq:featureenergy}
\end{equation}
where the expectation is taken over channels $c$ and horizontal positions $u$.
For each sample, we fit a line $\hat a z+\hat b$ to
$\log(e(z)+\epsilon)$ using least squares, with $\epsilon=10^{-8}$ for numerical
stability. The attenuation-consistency loss is defined as,\vspace{-0.5cm}
\begin{equation}
\Latt
=
\frac{1}{h}\sum_z
\left(
\hat a z+\hat b-\log\bigl(e(z)+\epsilon\bigr)
\right)^2 \vspace{-0.5cm}
\label{eq:atten}
\end{equation}
This regularizer encourages a smooth, approximately log-linear depth-wise energy
profile in the fused feature representation. The final training objective function is defined as,\vspace{-0.5cm}
\begin{equation}
\Lcal_{Total}
=
\Lfoc
+
\lambda_c\Lali
+
\lambda_a\Latt
\label{eq:total}
\end{equation}


\vspace{-0.7cm}
\section{Experiment}

\subsection{Datasets and preprocessing}
\label{sec:datasets}

We evaluate \textsc{PRCA-Net} on three public breast ultrasound (BUS) datasets:
BUSI~\cite{aldhabyani2020dataset}, BUS-BRA~\cite{gomezflores2024busbra}, and
UDIAT~\cite{yap2018automated}. Together, they provide variation in dataset size,
label provenance, and acquisition conditions. \textbf{BUSI} contains 780 images from 600
women, including 437 benign, 210 malignant, and 133 normal
images. We use the 647 lesion images for benign versus
malignant classification.
\textbf{BUS-BRA }provides the largest evaluation set, comprising 1,875 images from 1,064
patients acquired using four scanner models. Its labels are biopsy-proven, making
it particularly suitable for the main comparison. \textbf{UDIAT }contains 163 images
acquired with a Siemens ACUSON Sequoia C512 scanner, including 110 benign and 53
malignant lesions. BUS-BRA is split at the patient level, so that images from the same patient never appear in different partitions, whereas BUSI and UDIAT are split at the image level because their public releases provide no patient identifiers. For all datasets, we perform a class-stratified 70/15/15 train/validation/test split as described in Section~\ref{sec:protocol}; for BUS-BRA, the split is additionally grouped by patient. 


We perform splitting within each class so that the class distribution is preserved across the three
partitions. We apply the similar preprocessing pipeline to every
image: grayscale conversion, contrast-limited adaptive histogram equalization (CLAHE) with a clip limit of 2.0 and an $8\times8$ grid of local regions,
a $3\times3$ median filter, resizing to $224\times224$, and replication to three
channels for compatibility with ImageNet-pretrained backbones. We empirically select the values of hyper-parameters used in Eq. \ref{eq:total} to be $\lambda_c=\lambda_a=0.05$.
All values are means over five seeds on the held-out test partition, standard
deviations in parentheses, with Sensitivity and Specificity at the  best validation score.

\subsection{Training and evaluation protocol}
\label{sec:protocol}

Using a fixed random seed, we generate a class-stratified 70/15/15 train/validation/test split for each dataset. The split is implemented uniformly across all datasets as it is solely created from the class labels.We confirm that all of the available images are contained in the three disjoint divisions. While the validation set is used for threshold calibration, checkpoint selection, and early halting, the training set is utilized to update model parameters. The test
set is kept separate and is used only for the final evaluation.Using AdamW~\cite{loshchilov2019decoupled}, we train all models for 60 epochs with a weight decay of $10^{-4}$.  Using a step schedule with five-epoch intervals, the learning rate is lowered from $10^{-3}$ to $10^{-6}$. To prevent losing a significant portion of the smaller training set in each epoch, we employ a batch size of 16 for UDIAT and 32 for BUSI and BUS-BRA.  We use photometric and geometric augmentations during training. In order to maintain spatial consistency between the raw and restored streams of the same enhanced scan, the physics-restored image is created after augmentation. As the malignant form the minority class, a fixed decision threshold of
0.5 may produce biasness in the overall accuracy over sensitivity. We therefore select a decision threshold on the validation set by maximizing Youden's index~\cite{youden1950index}, and apply the selected threshold unchanged to the test set. 
\vspace{-0.5cm}
\begin{equation}
    \tau^\star = \arg\max_{\tau}
\left\{\mathrm{Sen}(\tau) + \mathrm{Spe}(\tau) - 1\right\},\vspace{-0.5cm}
\end{equation}

Each configuration is evaluated across five
runs with different random seeds; each run re-partitions the data and
re-initializes the model. Method comparisons use matched runs, allowing paired
$t$-tests over the five seed-wise results.

\subsection{Comparison with baselines}

In Table~\ref{tab:sota}, we compare our proposed \textsc{PRCA-Net} with state-of-the-art
methods on three BUS datasets, using results reported in the
respective studies. Our model achieves the highest accuracy, F1,
and AUROC among the listed methods on BUSI
(92.86\%, 91.94\%, and 97.10\%) and BUS-BRA
(93.24\%, 92.17\%, and 96.14\%). On UDIAT, its accuracy of
91.67\% is higher than those of Gheflati and Rivaz~\cite{gheflati2022vit}
and CTG-Net~\cite{yang2022ctgnet}, but lower than those of
CTMF-Net~\cite{wang2025ctmf} and HED-Net~\cite{koshy2025hednet}.
Its F1 of 89.92\% is higher than CTG-Net's but lower than
HED-Net's, while its AUROC of 82.81\% is higher only than that
of Gheflati and Rivaz. Our experiments use patient-disjoint splits
for BUS-BRA and image-level splits for BUSI and UDIAT, where
patient identifiers are unavailable.

\begin{table}[!tb]
\centering
\scriptsize
\setlength{\tabcolsep}{2pt}
\setlength{\aboverulesep}{1.5pt}
\setlength{\belowrulesep}{1.5pt}
\renewcommand{\arraystretch}{0.95}

\caption{Comparison of our proposed \textsc{PRCA-Net} against the state-of-the-art methods for benign and malignant BUS classification across three datasets.}
\label{tab:sota}
\begin{tabular}{cllccc}
\toprule
& \textbf{Method} & \textbf{Protocol} & \textbf{Acc} & \textbf{F1} & \textbf{AUROC} \\
\midrule
\multirow{3}{*}{\rotatebox[origin=c]{90}{\textit{BUS-BRA}}}
& Baseline~\cite{gomezflores2024busbra} & 1{,}875 images & 86.50 & --    & 93.10 \\
& SUA-Net~\cite{alhaj2026speckle}      & 1{,}875 images & 87.00 & 79.10 & 90.90 \\
& \textsc{PRCA-Net} (ours)             & 1{,}875 images & \textbf{93.24} & \textbf{92.17} & \textbf{96.14} \\
\midrule
\multirow{5}{*}{\rotatebox[origin=c]{90}{\textit{BUSI}}}
& HED-Net~\cite{koshy2025hednet}       & 647 images   & 88.46 & 88.45 & 95.38 \\
& CTG-Net~\cite{yang2022ctgnet}        & 647 images   & 90.00 & 85.00 & 89.00 \\
& MSIB~\cite{liu2026msib}              & 647 images   & 90.30 & 86.10 & 94.30 \\
& Byra~\cite{byra2021drs}              & 647 images   & 91.50 & --    & 95.50 \\
& \textsc{PRCA-Net} (ours)             & 647 images   & \textbf{92.86} & \textbf{91.94} & \textbf{97.10} \\
\midrule
\multirow{5}{*}{\rotatebox[origin=c]{90}{\textit{UDIAT}}}
& Gheflati \& Rivaz~\cite{gheflati2022vit} & 163 images   & 74.00 & --    & 80.00 \\
& CTG-Net~\cite{yang2022ctgnet}        & 163 images   & 86.00 & 80.00 & 86.00 \\
& CTMF-Net~\cite{wang2025ctmf}         & 163 images   & 92.04 & --    & 85.05 \\
& HED-Net~\cite{koshy2025hednet}       & 163 images   & \textbf{96.97} & \textbf{95.24} & \textbf{99.17} \\
& \textsc{PRCA-Net} (ours)             & 163 images   & 91.67 & 89.92 & 82.81 \\
\bottomrule
\end{tabular}
\end{table}

Along with this, Table~\ref{tab:encoder} compares different encoder backbones within the proposed framework. Overall, the proposed \textsc{PRCA-Net} achieves the most consistent performance across the three datasets. On BUS-BRA, it attains the best accuracy of 93.24\%, F$_1$-score of 92.17\%, specificity of 96.32\%, and AUC of 96.1, outperforming the strongest alternative, DenseNet-121, by 3.20 $\%$ and 3.61 $\%$ in accuracy and F$_1$, respectively, while improving AUC from 94.0 to 96.1. On BUSI dataset, \textsc{PRCA-Net} - achieves 92.86\% accuracy, 91.94\% F$_1$, and 97.1 AUC, giving the highest AUC among all evaluated backbones and matching the best F$_1$-score. On UDIAT, PRCA-Net achieves 91.67\% accuracy and 89.92\% F1, the highest among the evaluated backbones, and 100\% specificity. Its AUC of 82.81\%, however, is lower than that of all five backbones (84.4--90.6\%). Given the 24-image UDIAT test set (8 malignant), this AUC estimate is highly uncertain. These results suggest that the proposed architecture benefits from a stronger encoder while maintaining robustness across datasets, with the most pronounced gains observed on the BUS-BRA dataset.




\newcommand{\XS}{--.--{\tiny$\pm$-.-}}
\newcommand{\bXS}{\textbf{--.--}{\tiny$\pm$-.-}}

\begin{table}[!tbh]
\centering
\caption{Comparison of the proposed \textsc{PRCA-Net} against the SOTA image backbones on the three datasets.}
\label{tab:encoder}
\scalebox{0.95}{
\scriptsize
\begin{tabular}{l ccccc|ccccc| ccccc}
\toprule
& \multicolumn{5}{c|}{\textbf{BUS-BRA}}
& \multicolumn{5}{c|}{\textbf{BUSI}}
& \multicolumn{5}{c}{\textbf{UDIAT}} \\
\cmidrule(lr){2-6}\cmidrule(lr){7-11}\cmidrule(lr){12-16}
Backbone & Acc. & F$_1$ & Sen. & Spe. & AUC
         & Acc. & F$_1$ & Sen. & Spe. & AUC
         & Acc. & F$_1$ & Sen. & Spe. & AUC \\
\midrule
ResNet-18~\cite{he2016deep}
  & 87.54 & 85.90 & 82.42 & 90.00 & 94.0 & 91.84 & 90.86 & 90.62 & 92.42 & 95.10 & 83.33 & 77.78 & 50.00 & 100.00 & \textbf{90.60} \\
ResNet-34~\cite{he2016deep}
  & 89.68 & 88.38 & 86.81 & 91.05 & 92.70 & 90.82 & 89.29 & 81.25 & 95.45 & 94.70 & 79.17 & 78.22 & \textbf{87.50} & 75.00 & 89.10 \\
ResNet-50~\cite{he2016deep}
  & 82.92 & 81.45 & 84.62 & 82.11 & 88.70 & 92.86 & 91.94 & 90.62 & 93.94 & 92.30 & 87.50 & 85.45 & 75.00 & 93.75 & 87.50 \\
DenseNet-121~\cite{huang2017densely}
  & 90.04 & 88.56 & 83.52 & 93.16 & 93.80 & 90.82 & 89.64 & 87.50 & 92.42 & 93.80 & 87.50 & 84.18 & 62.50 & 100.00 & 84.40 \\
MobileNetV2~\cite{sandler2018mobilenetv2}
  & 74.73 & 70.35 & 56.04 & 83.68 & 75.70 & 88.78 & 87.53 & 87.50 & 89.39 & 95.70 & 87.50 & 84.18 & 62.50 & 100.00 & 88.30 \\
\midrule
\textbf{\textsc{PRCA-Net} (Ours)}
  & \textbf{93.24}
  & \textbf{92.17} & \textbf{86.81} & \textbf{96.32} & \textbf{96.14 }
  & \textbf{92.86} & \textbf{91.94} & \textbf{90.62} & \textbf{93.94} & \textbf{97.10} & \textbf{91.67} & \textbf{89.92} & 75.00 & \textbf{100.00} & \textbf{82.81} \\
\bottomrule
\end{tabular}}
\end{table}

\subsection{Ablation studies}

\noindent\textbf{Architecture ablation:} Table~\ref{tab:stream_ablation} presents an architectural ablation of \textsc{PRCA-Net} by progressively introducing the restored stream and the proposed fusion mechanism. On BUS-BRA, the two-stream concatenation improves accuracy and F$_1$ by 1.42 $\%$ and 1.34 $\%$, respectively, over the raw-only baseline. Replacing simple concatenation with the proposed \textsc{PRCA-Net}  further increases accuracy and F$_1$ to 93.24\% and 92.17\%, corresponding to additional gains of 1.43 $\%$ and 1.71 $\%$, while achieving an AUC of 96.14 $\%$. On BUSI, the two-stream model substantially improves over the raw-only model, increasing accuracy and F$_1$ by 3.06 $\%$ and 3.54 $\%$, respectively. \textsc{PRCA-Net}  maintains this strong performance while further improving AUC to 97.16 $\%$, compared with 96.12\% for simple concatenation. The largest accuracy and F1 gain is observed on UDIAT, where PRCA-Net improves over the two-stream concatenation model by 8.34\% and 10.09\%, respectively. Its UDIAT AUC (82.81\%) is above concatenation (78.91\%) but below the single-stream models (83.59\% and 89.84\%); given the 8 malignant test images, we do not draw conclusions from UDIAT AUC. Overall, these results demonstrate that the performance gains arise not simply from introducing a second representation, but from the proposed architecture for effectively integrating the raw and restored views.

\vspace{-0.3cm}
\begin{table*}[!tbh]
\centering
\caption{Ablation study of the dual-stream architecture on BUSBRA, BUSI, and UDIAT. All values are reported in percentages (\%).}
\label{tab:stream_ablation}
\resizebox{\textwidth}{!}{%
\begin{tabular}{lccccc|ccccc|ccccc}
\hline
\textbf{Variant}
& \multicolumn{5}{c|}{\textbf{BUSBRA}}
& \multicolumn{5}{c|}{\textbf{BUSI}}
& \multicolumn{5}{c}{\textbf{UDIAT}} \\
& Acc. & F1 & Sen. & Spec. & AUC
& Acc. & F1 & Sen. & Spec. & AUC
& Acc. & F1 & Sen. & Spec. & AUC \\
\hline

Raw-only
& 90.39 & 89.12 & 86.81 & 92.11 & 94.60
& 89.80 & 88.40 & 84.38 & 92.42 & 93.42
& 87.50 & 85.45 & 75.00 & 93.75 & \textbf{83.59} \\

Restored-only
& 89.68 & 88.11 & 82.42 & 93.16 & 95.02
& 90.82 & 89.80 & 90.62 & 90.91 & 94.55
& 87.50 & 84.18 & 62.50 & 100.00 & 89.84 \\

Two-stream (Concat.)
& 91.81 & 90.46 & 83.52 & 95.79 & 94.90
& 92.86 & 91.94 & 90.62 & 93.94 & 96.12
& 83.33 & 79.83 & 62.50 & 93.70 & 78.91 \\

\textbf{\textsc{PRCA-Net} }
& \textbf{93.24} & \textbf{92.17} & \textbf{86.81} & \textbf{96.32} & \textbf{96.14}
& \textbf{92.86} & \textbf{91.94} & \textbf{90.62} & \textbf{93.94} & \textbf{97.16}
& \textbf{91.67} & \textbf{89.92} & \textbf{75.00} & \textbf{100.00} & 82.81 \\

\hline
\end{tabular}}
\end{table*}

\begin{figure}[!tbh]
\centering
\includegraphics[width=0.7\linewidth]{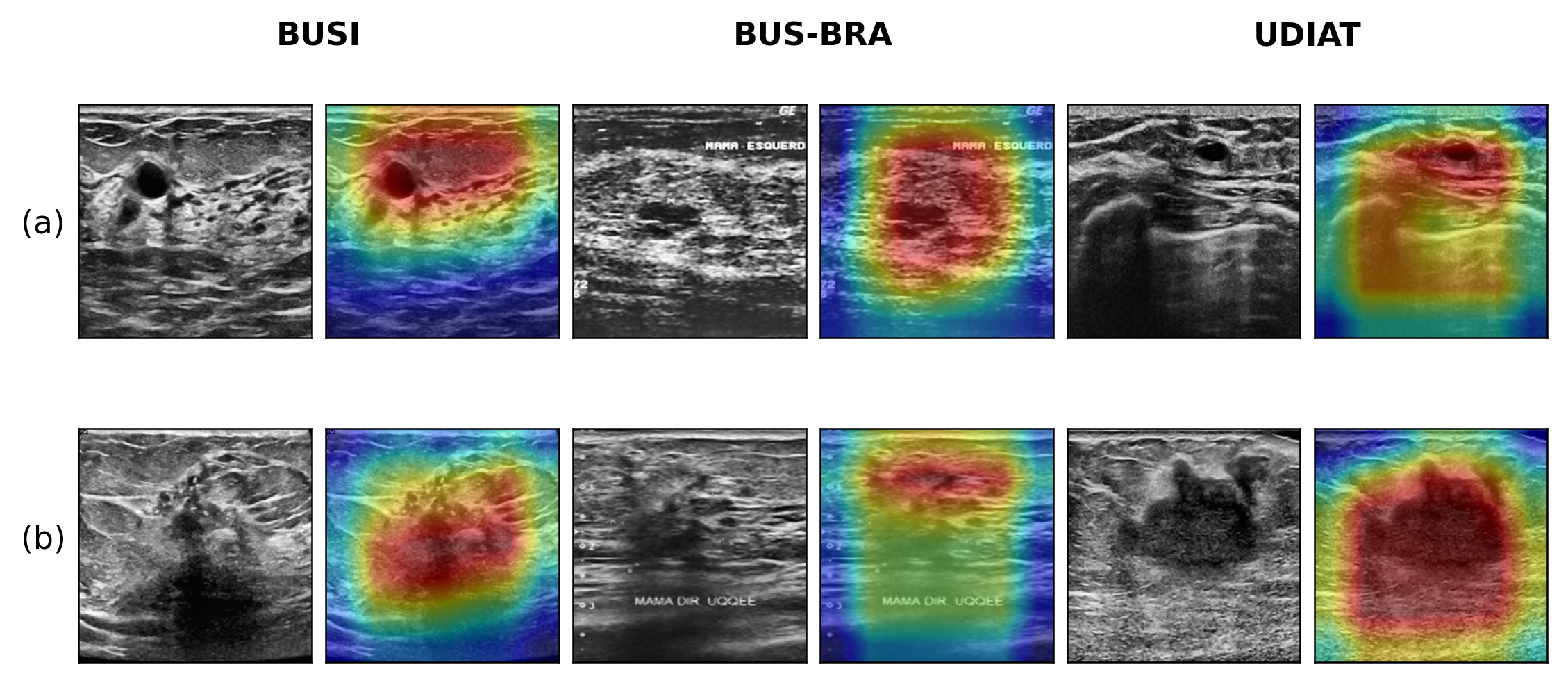}
\caption{Grad-CAM for \textsc{PRCA-Net}  for the
BUSI, BUS-BRA and UDIAT datasets. Rows (a) and (b) depict a benign and a malignant case respectively.}
\label{fig:gradcam}
\end{figure}

\noindent\textbf{Qualitative analysis:} In Figure~\ref{fig:gradcam}, we show the Grad-CAM visualizations for BUSI,
BUS-BRA, and UDIAT dataset test images. In each pair, the left image is the B-mode scan and the right image is the corresponding activation map; rows (a) and (b) show two examples with different lesion appearances. Across the three datasets, the model mainly attends to the lesion and its nearby tissue context. These maps suggest that the model uses lesion morphology together with local acoustic context rather than relying on intensity alone.

\noindent\textbf{Effect of loss components:} In Table~\ref{tab:loss_ablation}, we evaluate the contribution of the proposed alignment
and attenuation-consistency losses. On BUS-BRA and BUSI, the complete objective
achieves the best overall results, improving accuracy by 5.70\% and 4.08\% over
the focal-loss baseline, respectively. It also yields the highest F1-score and
specificity on both datasets, indicating that the two regularizers provide
complementary benefits when used together.

The individual loss terms also improve selected metrics. The alignment loss
notably increases specificity on BUS-BRA, while the attenuation loss improves
sensitivity and AUC on BUSI. On the smaller UDIAT dataset, the results are more
variable: the complete objective matches the focal-loss baseline in accuracy and
improves specificity to 100\%, but focal loss alone achieves higher sensitivity
and AUC. This variation is expected given the limited test set size. Overall, the
results on the larger BUS-BRA and BUSI datasets support combining alignment and
attenuation constraints with the classification objective.

\begin{table*}[!tbh]
\centering
\renewcommand{\arraystretch}{1.3}
\setlength{\tabcolsep}{4pt}
\caption{Ablation study of loss configurations on BUS-BRA, BUSI, and UDIAT.
All values are reported in percentage (\%). Bold indicates the best result for
each metric within a dataset; tied best values are both highlighted.}
\label{tab:loss_ablation}
\resizebox{\textwidth}{!}{%
\begin{tabular}{lccccc|ccccc|ccccc}
\hline
\textbf{Loss configuration}
& \multicolumn{5}{c|}{\textbf{BUS-BRA}}
& \multicolumn{5}{c|}{\textbf{BUSI}}
& \multicolumn{5}{c}{\textbf{UDIAT}} \\
& Acc. & F1 & Sen. & Spec. & AUC
& Acc. & F1 & Sen. & Spec. & AUC
& Acc. & F1 & Sen. & Spec. & AUC \\
\hline

$\mathcal{L}_{\mathrm{focal}}$
& 87.54 & 85.90 & 82.42 & 90.00 & 93.07
& 88.78 & 87.34 & 84.38 & 90.91 & 94.32
& \textbf{91.67} & \textbf{90.62} & \textbf{87.50} & 93.75 & \textbf{95.31} \\

$\mathcal{L}_{\mathrm{focal}}+
\lambda_c\mathcal{L}_{\mathrm{align}}$
& 91.10 & 89.63 & 82.42 & 95.26 & 94.68
& 89.80 & 88.58 & 87.50 & 90.91 & 93.84
& \textbf{91.67} & 89.92 & 75.00 & \textbf{100.00} & 85.16 \\

$\mathcal{L}_{\mathrm{focal}}+
\lambda_a\mathcal{L}_{\mathrm{atten}}$
& 91.81 & 90.73 & \textbf{89.01} & 93.16 & \textbf{96.54}
& 88.78 & 87.70 & \textbf{90.62} & 87.88 & 95.41
& 87.50 & 86.34 & \textbf{87.50} & 87.50 & 88.28 \\

\textbf{$\mathcal{L}_{\mathrm{focal}}+
\lambda_c\mathcal{L}_{\mathrm{align}}+
\lambda_a\mathcal{L}_{\mathrm{atten}}$}
& \textbf{93.24} & \textbf{92.17} & 86.81 & \textbf{96.32} & 96.14
& \textbf{92.86} & \textbf{91.94} & \textbf{90.62} & \textbf{93.94} & \textbf{97.16}
& \textbf{91.67} & 89.92 & 75.00 & \textbf{100.00} & 82.81 \\

\hline
\end{tabular}}
\end{table*}

\begin{table*}[!tbh]
\centering
\renewcommand{\arraystretch}{1.3}
\setlength{\tabcolsep}{4pt}
\caption{Ablation study of feature-alignment strategies on BUS-BRA, BUSI, and
UDIAT. All values are reported in percentage (\%). Bold indicates the best result
for each metric within a dataset; tied best values are both highlighted.}
\label{tab:alignment_ablation}
\resizebox{\textwidth}{!}{%
\begin{tabular}{lccccc|ccccc|ccccc}
\hline
\textbf{Alignment strategy}
& \multicolumn{5}{c|}{\textbf{BUS-BRA}}
& \multicolumn{5}{c|}{\textbf{BUSI}}
& \multicolumn{5}{c}{\textbf{UDIAT}} \\
& Acc. & F1 & Sen. & Spec. & AUC
& Acc. & F1 & Sen. & Spec. & AUC
& Acc. & F1 & Sen. & Spec. & AUC \\
\hline

No alignment
& 91.10 & 89.69 & 83.52 & 94.74 & 94.29
& 90.82 & 89.29 & 81.25 & \textbf{95.45} & 90.48
& 83.33 & 79.83 & 62.50 & 93.75 & 82.03 \\

Standard cosine alignment
& 91.10 & 89.69 & 83.52 & 94.74 & 94.03
& 91.84 & 90.72 & 87.50 & 93.94 & 93.80
& 83.33 & 81.25 & \textbf{75.00} & 87.50 & \textbf{85.94} \\

\textbf{Banded cosine alignment}
& \textbf{93.24} & \textbf{92.17} & \textbf{86.81} & \textbf{96.32} & \textbf{96.14}
& \textbf{92.86} & \textbf{91.94} & \textbf{90.62} & 93.94 & \textbf{97.16}
& \textbf{91.67} & \textbf{89.92} & \textbf{75.00} & \textbf{100.00} & 82.81 \\

\hline
\end{tabular}}
\end{table*}

\noindent\textbf{Effect of feature alignment.}
Table~\ref{tab:alignment_ablation} compares no alignment, standard cosine
alignment, and the proposed banded cosine alignment. On the BUSI dataset, the banded cosine alignment obtains the highest accuracy, F1-score, sensitivity, and AUC, while no alignment yields the highest
specificity. In relative to no alignment, the banded alignment improves accuracy by
2.14\% on BUS-BRA and 2.04\% on BUSI dataset, with sensitivity gains of 3.29\% and
9.37\%, respectively. On the UDIAT dataset, it improves accuracy by 8.34\%, F1-score by
10.09\%, and specificity by 6.25\% over no alignment, while matching the highest
sensitivity of 75\%. The standard cosine alignment improves several metrics on the BUSI dataset over no alignment, but yields less consistent results across the three datasets. Overall, the results demonstrate that maintaining a bounded similarity between the two streams rather than directly maximizing their agreement.
\vspace{-0.3cm}
\section{Conclusion}
\label{sec:conclusion}

We presented \textsc{PRCA-Net}, a dual-stream framework for breast ultrasound
classification that combines physics-augmented and physics-restored views through
bidirectional cross-attention. By retaining both views, the model can use
restoration-derived information without discarding acquisition-dependent cues that
may remain diagnostically useful. The proposed banded cosine alignment objective
encourages correspondence between the two streams while preserving view-specific
information, and the attenuation-consistency loss incorporates an
ultrasound-inspired depth-wise prior into the fused representation.   The experiments conducted on the BUSI, BUS-BRA, and UDIAT datasets show that \textsc{PRCA-Net} offers comparable classification performance across datasets with various image features. While Grad-CAM images demonstrate that the model mainly considers lesion sites and their surrounding tissue context, the ablations validate the importance of limited feature alignment and the attenuation-consistency target. These
results suggest that explicitly modeling complementary physics-guided views can be
useful for BUS classification. The present study is limited to public B-mode datasets with image-level labels, and BUSI and UDIAT are split at the image level, so images from the same patient may appear in different partitions. Future work will evaluate the framework on larger patient-disjoint and multi-institutional cohorts, and extend it to radio-frequency ultrasound data, where acquisition effects can be modeled more directly.









\vspace{-0.3cm}
\bibliographystyle{elsarticle-harv}
\bibliography{references.bib}

\end{document}